\documentclass[sigconf,screen]{acmart}

\AtBeginDocument{%
  \providecommand\BibTeX{{%
    \normalfont B\kern-0.5em{\scshape i\kern-0.25em b}\kern-0.8em\TeX}}}

\setcopyright{acmlicensed}
\copyrightyear{2018}
\acmYear{2018}
\acmDOI{XXXXXXX.XXXXXXX}

\acmConference[Conference acronym 'XX]{Make sure to enter the correct
  conference title from your rights confirmation emai}{June 03--05,
  2018}{Woodstock, NY}
\acmISBN{978-1-4503-XXXX-X/18/06}

\usepackage{graphicx}
\usepackage{subfigure}
\usepackage{enumitem}
\usepackage{makecell}
\newcommand{\vpara}[1]{\vspace{1.5ex}\noindent\textbf{#1}}

\newcommand{\model}{{KBLLaMA}}
\newcommand{\smodel}{{KBLLaMA} }

\usepackage{multirow}
\usepackage{algorithm}
\usepackage{algorithmic}
\usepackage{appendix}
\usepackage{titletoc}
\begin{document}
\title{\textsc{\model}: A Learn-Then-Reason Model Towards Generalization in Knowledge Base Question Answering}

\author{Lingxi Zhang}
\affiliation{%
  \institution{Renmin University of China}
  \city{Beijing}
  \country{China}}
\email{zhanglingxi@ruc.edu.cn}

\author{Jing Zhang}
\affiliation{%
  \institution{Renmin University of China}
  \city{Beijing}
  \country{China}}
\email{zhang-jing@ruc.edu.cn}

\author{Yanling Wang}
\affiliation{%
  \institution{Zhongguancun Laboratory}
  \city{Beijing}
  \country{China}}
\email{wangyl@zgclab.edu.cn}

\author{Cuiping Li}
\affiliation{%
  \institution{Renmin University of China}
  \city{Beijing}
  \country{China}}
\email{licuiping@ruc.edu.cn}

\author{Hong Chen}
\affiliation{%
  \institution{Renmin University of China}
  \city{Beijing}
  \country{China}}
\email{chong@ruc.edu.cn}



\begin{abstract}
Large-scale knowledge bases (KBs) like Freebase and Wikidata house millions of structured knowledge. Knowledge Base Question Answering (KBQA) provides a user-friendly way to access these valuable KBs via asking natural language questions.
In order to improve the generalization capabilities of KBQA models, extensive research has embraced a retrieve-then-reason framework to retrieve relevant evidence for logical expression generation.
These multi-stage efforts prioritize acquiring external sources but overlook the incorporation of new knowledge into their model parameters.
In effect, even advanced language models and retrievers have knowledge boundaries, thereby limiting the generalization capabilities of previous KBQA models.
Therefore, this paper develops \model, which follows a learn-then-reason framework to inject new KB knowledge into a large language model for flexible end-to-end KBQA. At the core of \model, we study (1) how to organize new knowledge about KBQA and (2) how to facilitate the learning of the organized knowledge.
Extensive experiments on various KBQA generalization tasks showcase the state-of-the-art performance of \model. Especially on the general benchmark GrailQA and domain-specific benchmark Bio-chemical, \smodel respectively derives a performance gain of up to 3.8\% and 9.8\% compared to the baselines..
\end{abstract}




\maketitle

\section{Introduction}
\begin{figure}[t]
	\centering
	\subfigure[Examples of In-KB and Cross-KB generalization in KBQA.] { 
		\includegraphics[width=3.3 in]{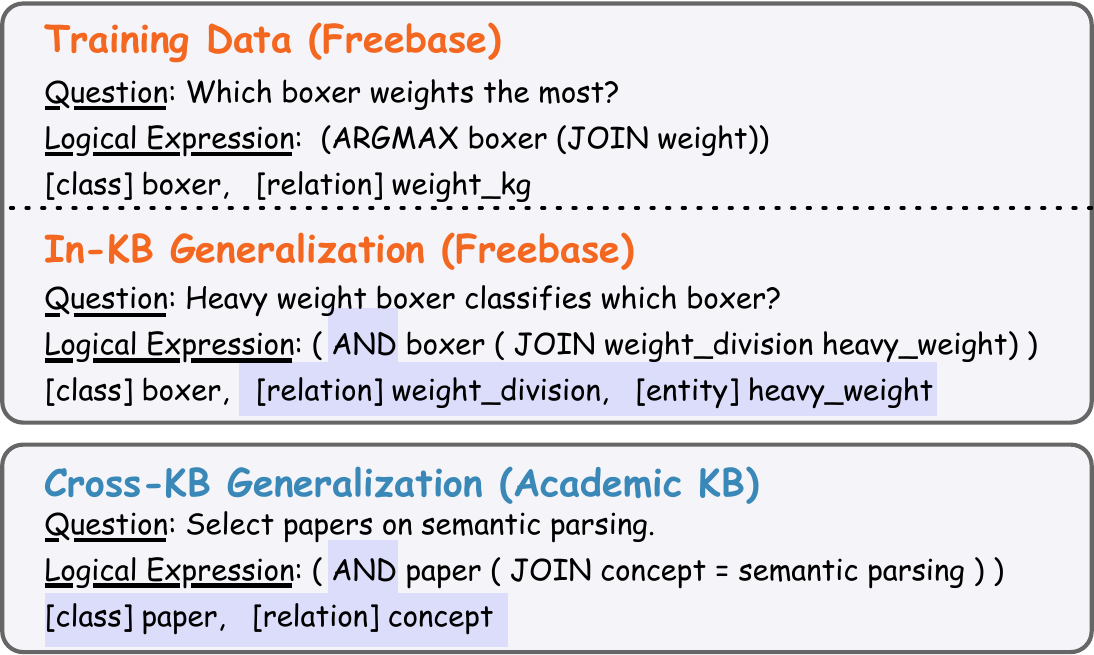}
		\label{fig:generalization_examples}
	}
	\hspace{0.05in}
        \centering
	\subfigure[Retrieve-Then-Reason vs. Learn-Then-Reason.] { 
		\includegraphics[width=3.3in]{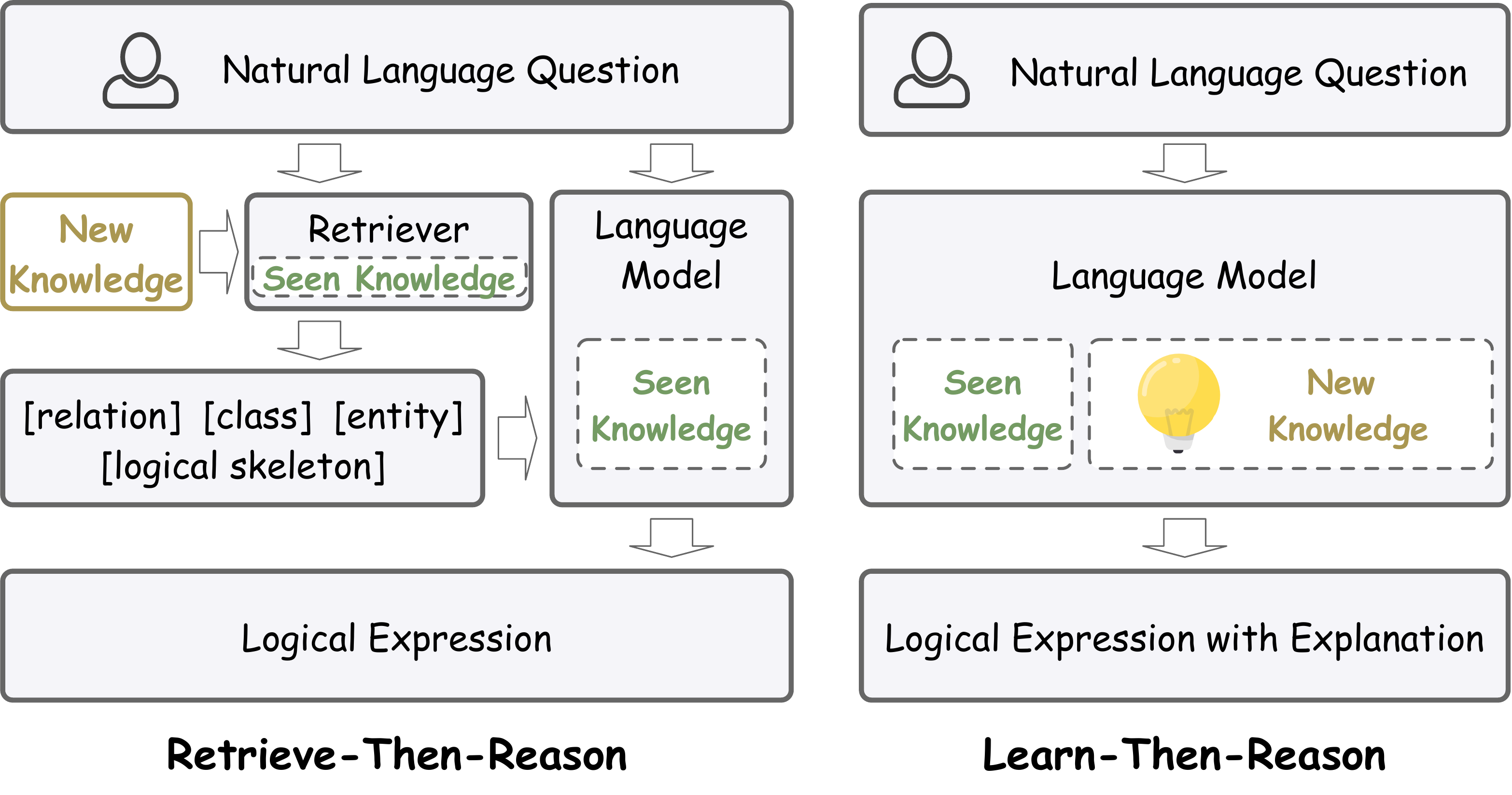} 
		\label{fig:RetrieveVslearn}
	}
	\hspace{0.05in}
	
	\caption{Approaches towards generalization in KBQA. (a) Examples of In-KB and Cross-KB generalization. The new KB knowledge is colored in purple. (b) Comparison between the traditional retrieve-then-reason approach and our proposed learn-then-reason approach. The former overlooks the incorporation of new KB knowledge into their model parameters, thereby limiting their generalization capabilities.}
	\label{fig:intro}
\end{figure}

Currently, the most advanced methods of KBQA are based on semantic parsing (SP)~\cite{luo2023chatkbqa,zhang2023fc-kbqa,ye2022rng,gu-etal-2023-dont}, which convert natural language questions into logical expressions automatically and execute them on KBs to obtain answers.
Most SP-based methods~\cite{das2021case,lan2020query,yih2016value} have demonstrated success in the Independent and Identically Distributed (I.I.D.) setting, where the logical expressions in both training set and test set are based to the same KB and share the same distribution.
Nevertheless, in real-world scenarios, In-KB generalization and Cross-KB generalization illustrated in Figure~\ref{fig:generalization_examples} are more commonly encountered.
For the In-KB setting, the training set and test set are based on the same KB but exhibit distinct distributions. Precisely, the test set includes new compositions of seen relations and classes in the training set, or even involves new relations and classes.
The Cross-KB setting is a more intricate scenario wherein the training set and test set originate from different KBs, featuring diverse ontology formats and even distinct domain knowledge.


Numerous researches have benefited from pre-trained language models (PLMs) like BERT~\cite{devlin2018bert}, BART~\cite{lewis2020bart}, and T5~\cite{2020t5} to enhance KBQA models' generalization capabilities.
Typically, 
retriever-based models like FC-KBQA~\cite{zhang2023fc-kbqa} employ retrievers to identify the most relevant relations and classes to the question. Afterwards, the retrieved elements are fed into a PLM alongside the question to produce the logical expressions. 
Recently, motivated by the remarkable success of large language models (LLMs), some approaches feed the retrieved elements into an LLM for generating logical expressions~\cite{jiang-etal-2023-structgpt,wang2023keqing,wang2023knowledgedriven}. The extensive model parameters and vast training corpora empower LLMs to acquire substantial knowledge and exhibit promising capabilities in knowledge reasoning.
However, these \textbf{retrieve-then-reason} methods are hindered by their complex multi-module framework. More importantly, even advanced LLMs and retrievers have knowledge boundaries, especially within specific domains like finance and medicine. Existing methods overlook the incorporation of new/unseen KB knowledge into their model parameters, thereby limiting their generalization capabilities, especially in the Cross-KB setting.

To address the problem, we propose a \textbf{learn-then-reason} approach, aiming to incorporate new KB knowledge into an LLM' model parameters for the end-to-end KBQA (Cf. Figure~\ref{fig:RetrieveVslearn}).
More specifically, we organize the new knowledge about the KBQA task as <question, logical\_expression> training pairs, which are subsequently used to fine-tune the popular open-source LLaMA2-7B~\cite{touvron2023llama} to derive our KBQA model named \model. 
In practice, KBLLaMA-base is trained on general KBQA benchmark datasets and augmented training data from popular general KBs, aiming to capture general knowledge for In-KB and Cross-KB generalization.
Users can customize \smodel for domain-specific KBs via continuously fine-tuning the \model-Base with knowledge derived from the domain-specific KBs, enhancing the Cross-KB generalization capability.
To achieve the above goal, two research questions are raised: \textbf{RQ1}: How to organize the new KB knowledge? and \textbf{RQ2}: How to facilitate the learning of the organized knowledge? 

In this paper, we propose to organize the new KB knowledge into a number of diverse and high-quality <question, logical\_expression> training pairs (RQ1). Specifically, we feed KB-constrained prompts that express a logical expression into the powerful GPT-3.5\footnote{\url{https://openai.com/blog/chatgpt}} to generate high-quality natural language questions.
To create the logical expressions, a straightforward method is to enumerate all possible logical expressions involved in a KB. However, the enumeration process is computationally inefficient for complex multi-hop logical expressions. 
Current studies on LLM fine-tuning have highlighted the importance of ``less is more'' ~\cite{zhou2023lima}, i.e., improving the diversity and quality of training data is more important than increasing data quantity. Therefore, we design a cluster-based strategy to select a subset of complex logical expressions, ensuring that the selected logical expressions exhibit high diversity and good quality.
Then we use the selected complex logical expressions as well as all the one-hop logical expressions for training pair construction.
Furthermore, considering that the main target of LLaMA2 is to comprehend natural languages rather than logical expressions, we refine the initial augmented training pairs to facilitate the learning process (RQ2). The knowledge refinement process is inspired by the chain-of-thought (CoT)~\cite{wei2022chain} to incorporate the name of the queried KB and a natural language explanation of the logical expression before the returned logical expression.

To sum up, our contributions are summarized as follows:

\begin{itemize}[leftmargin=1em]
    \item
     We propose a learn-then-reason approach for KBQA, which incorporates new KB knowledge into the model parameters to fundamentally improve the model's generalization capabilities. Moreover, by eliminating the requirement of external retrievers, our approach facilitates more flexible end-to-end reasoning.
 
 \item We propose a KB-aware data construction strategy for determining and organizing the new knowledge to be learned, ensuring both the diversity and quality of the generated data while maintaining a balanced trade-off between efficiency and effectiveness.
    \item 
 Extensive experiments demonstrate the promising performance of \smodel for both In-KB and Cross-KB generalization tasks. Especially on the general benchmark GrailQA and domain-specific benchmark Bio-chemical, \smodel respectively derives a performance gain of up to 3.8\% and 9.8\% compared to other KBQA baselines and the most capable commercial LLM GPT-4.
\end{itemize}

\section{Related Work}
Numerous studies have been focused on generalization issues of KBQA, and can categorize into traditional KBQA methods and LLM-based KBQA methods according to whether LLMs are utilized.

\subsection{Traditional Semantic Parsing-Based KBQA}
SP-based KBQA is currently the most effective and popular KBQA framework, which first translates questions into logical expressions and then yield answers by executing the logical expressions on KBs.
To adapt to out-of-distribution questions,
some SP-based approaches, such as GrailQA-Rank~\cite{gu2021beyond} and QGG~\cite{lan2020query}, employ a PLM-based rank module for semantic matching between the question and the logical expression.
These approaches enumerate logical expression candidates or query graphs~\cite{lan2020query} (a query graph can be transformed into a logical expression) associated with the KB and rank them according to their relevance to the given question.
In addition to rank-based models, another line of SP-based approaches produce logical expressions via a generation model.
These approaches employ a rank-based model as a retriever to acquire relevant KB information, which is then utilized as augmentation to be fed into the generation model.
For instance, RNG-KBQA~\cite{ye2022rng} adopts top-k ranked candidate logical expressions as the augmentation. Some methods~\cite{zhang2023fc-kbqa,shu2022tiara} propose a more fine-grained augmentation by retrieving KB components like relations, classes, and logical skeletons. Specially, ChatKBQA~\cite{luo2023chatkbqa} adopts a generate-then-retrieve framework, which first fine-tunes the LLaMA model to generate raw logical expression candidates and then retrieves relation candidates to refine the logical expressions. 
While the aforementioned methods can mitigate the generalization problem in KBQA, they do not
embed new knowledge about the KBQA task into the model parameters, thereby restricting their capabilities, particularly in the Cross-KB setting.
This work suggests another view by injecting new KB knowledge into the model parameters to enhance the model’s performance in both In-KB and Cross-KB settings.


\subsection{LLM-Based KBQA}
To expand the knowledge of KBQA models, LLMs such as ChatGPT or LLaMA are leveraged.
For instance, ~\citet{tan2023chatgpt} directly employ GPT-3.5 to generate answers in response to natural questions and incorporate an answer matching strategy to enhance the prediction accuracy. 
To deal with more complex questions, StructGPT~\cite{jiang-etal-2023-structgpt} constructs specialized interfaces to retrieve relevant evidence from structured data such as KB.
Considering that LLMs perform well on code generation tasks, KB-coder~\cite{nie2024codestyle} introduces a code-style in-context learning method, transforming the KB logical expression generation process into a Python-like code generation process. 
Apart from above training-free models, some approaches decouple the complex question reasoning into multiple steps and utilize trained retrievers to enhance each step. 
For instance, KD-CoT~\cite{wang2023knowledgedriven} decouples the questions via formulating a CoT rationale process to implement a retriever-reader-verifier framework.
Alternatively, Keqing~\cite{wang2023keqing} decomposes the complex questions based on predefined templates, retrieving candidate entities for each sub-questions and ultimately generating a response with reasoning paths.
However, these employed retrievers work in an I.I.D. setting, thereby facing generalization issues when encountering questions of unseen domains.
Besides directly using LLMs to conduct prediction like the above methods, FlexKBQA~\cite{li2023flexkbqa} distills knowledge from LLMs to further optimize the traditional KBQA model RNG-KBQA~\cite{ye2022rng}.
Despite the vast amount of knowledge embedded in model parameters, LLMs still have knowledge boundaries. Currently, LLM-based KBQA methods primarily focus on how to fully utilize the inherent knowledge and strong reasoning abilities of LLMs, neglecting the knowledge gap between LLMs and the specific KBQA tasks.
Therefore, our work is dedicated to bridging this knowledge gap.


\section{Preliminaries}
In this section, we first present the background of KBs. Then we introduce the basic pipeline SP-based KBQA approach and the definition of logical expression.

\subsection{Knowledge Base}
We denote $\mathcal{K}=\{{KB}_{1}, {K B}_2,...,{KB}_N\}$ as a set of knowledge bases, where each knowledge base $KB_i\in\mathcal{K}$ consists of an ontology $\mathcal{O}_i=\{(\mathcal{C}_i \times \mathcal{R}_i \times \mathcal{C}_i)\}$ and a set of relational facts $\mathcal{F}_i=\{(\mathcal{E}_i \times \mathcal{R}_i \times (\mathcal{E}_i\cup \mathcal{C}_i))\}$. Here $\mathcal{C}_i, \mathcal{R}_i,$ and $\mathcal{E}_i$ denote the set of classes, relations, and entities in $KB_i$, respectively. Specially, literals are treated as a special type of entity.
In practice, an entity in a logical expression is generally expressed by an entity ID, and each entity mention can correspond to multiple entity IDs.
A relational fact is represented as a <subject\_entity, relation, object\_entity> triplet, where the class that the subject entity belongs to is named domain class, and the class that the object entity belongs to is named range class.

The ontology defines the meta-information of a KB.
For example, <\textit{music.ablum}, \textit{music.album.album}\_\textit{artist}, \textit{music.artist}> is an ontology triplet that depicts an artist releasing an album, where ``\textit{music.ablum}'' and ``\textit{music.artist}'' are the domain class and range class of the relation ``\textit{music.album.album}'', respectively.
A fact triplet is an instance of an ontology triplet. For instance, <\textit{Opera Arias}, \textit{music.album.album}\_\textit{artist}, \textit{Samuel Rame}> is an instance of the above ontology triplet.
It is worth noting that distinct KBs possess different ontologies. Even within the same domain, KBs can have diverse ontologies.
For example, a relation that means ``the artist of a music album'' is termed ``\textit{music.album.artist}'' in Freebase while it is expressed as         ``\textit{artist}\_\textit{of}\_\textit{album}'' in Wikidata.



\subsection{SP-Based KBQA and Logical Expression}
Given a natural language question $q$ and a knowledge base ${KB}_i$, a KBQA model aims to find a set of entities denoted by $\mathcal{A}
\subseteq \mathcal{E}_i$ from ${KB}_i$ as the answer set to $q$.
Currently, the most effective KBQA approach is SP-based KBQA, which transforms $q$ to an executable logic expression $s$ and then executes $s$ on the KB to yield answers.
The logic expressions can be expressed by SPARQL, lambda-DCS~\cite{liang2013learning}, query graph~\cite{lan2020query}, and s-expression~\cite{gu2021beyond}. 
We choose the s-expression in this work because it more resembles natural language and demonstrates robust compositional capabilities. Although a majority of KBs require the SPARQL queries, s-expressions can be easily transformed to SPARQL to support the KBQA. The s-expression integrates relations, classes, entities, and a variety of logical operators. We denote a logical operator as $g$. Appendix~\ref{sec:appendix_a} introduces a variety of logical operators in s-expressions. 

\section{\model}
\label{sec:approach}
\begin{figure*}
\includegraphics[width=1\textwidth]{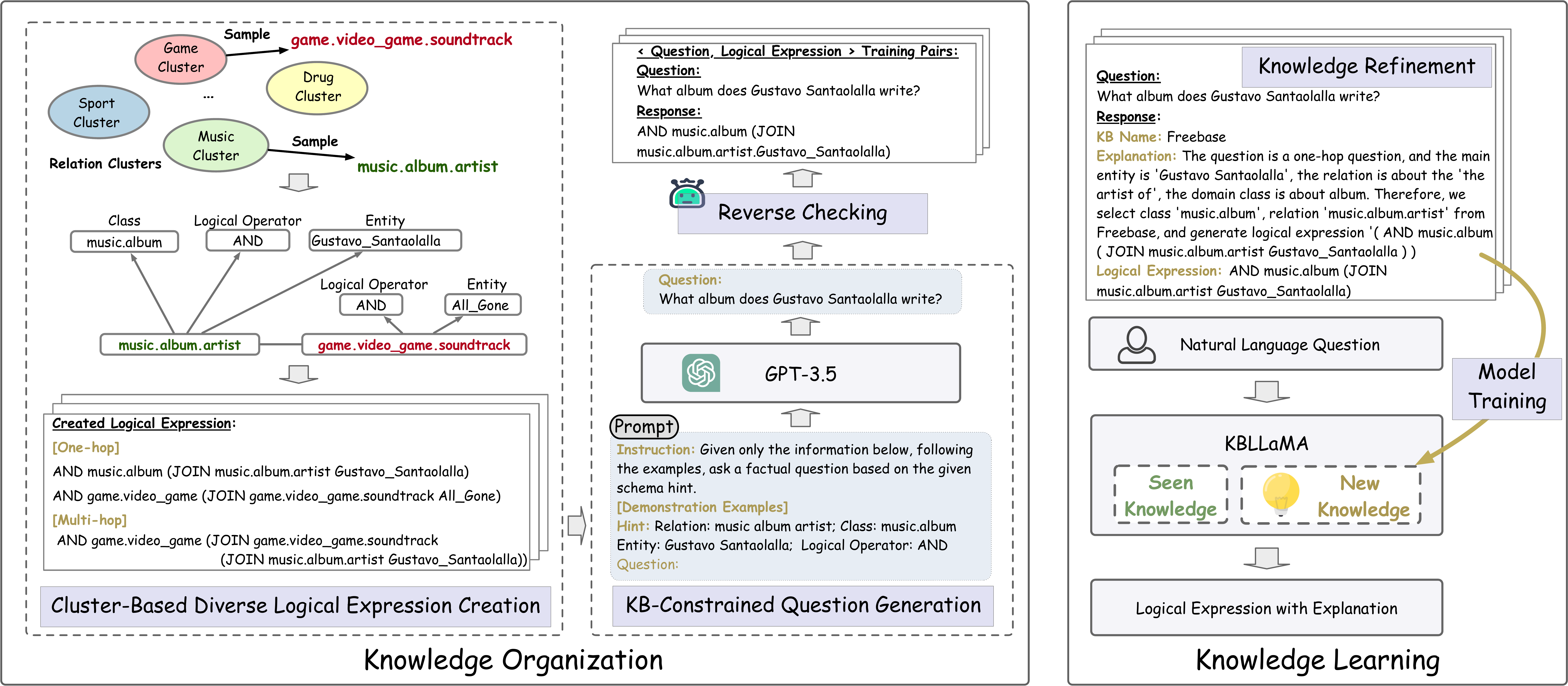} 
\caption{The overview of the development of \model.}
\label{fig: overview_development} 
\end{figure*}
In this section, we will  introduce our learn-then-reason model named \smodel for KBQA, explaining how to make \smodel own generalization capabilities to deal with unseen KB knowledge.

\subsection{Overview}
Distinct from traditional retrieve-then-reason KBQA models that rely on  semantic matching-based retrievers, our end-to-end method \smodel is built by fine-tuning the open-source LLaMA2-7B to incorporate new KB knowledge into model parameters for the end-to-end KBQA.
Therefore, how to organize the new knowledge and enable \smodel to learn from it becomes the key focus of our research.
Figure~\ref{fig: overview_development} illustrates the overview of  \model.
\textbf{Knowledge Organization}. 
We structure the unseen knowledge about KBQA as <question, logical\_expression> pairs to augment our training data. 
To do this,
we create multiple logical expressions and feed their key elements into GPT-3.5 to generate corresponding questions through KB-constrained in-context learning.
The augmented data is desired to involve all relations within the queried KB. However, enumerating all possible logical expressions about the queried KB is computationally in-efficient.
Recent studies~\cite{zhou2023lima} have highlighted that improving the diversity and quality of training data is more important than solely focusing on increasing data quantity.
Therefore, to ensure the diversity of the augmented data without exhaustively enumerating all logical expressions, we design a cluster-based strategy for creating a set of diverse logical expressions. Moreover, we introduce a reverse checking strategy to eliminate low-quality generated questions, enhancing the quality of our augmented data.

\textbf{Knowledge Learning}. 
We utilize the organized new knowledge, i.e., the augmented <question, logical\_expression> training pairs to fine-tune the open-source LLaMA2-7B to derive the \model.
Particularly, in order to facilitate the learning of \model, we follow the idea of chain-of-thought~\cite{wei2022chain} to refine the augmented <question, logical\_expression> pairs by including the name of the queried KB and the explanation of the logical expression.
This refinement process produces <question, KB\_name; explanation; logical\_expression> pairs.
Specially, \smodel is trained to generate entity mentions instead of the entity IDs which are numerical values without semantic information and is not conducive to the learning of \model.
Therefore, we employ a trainable entity linker to match entity mentions to specific entity IDs.

\subsection{Knowledge Organization}
The automated knowledge organization procedure consists of the following three steps to obtain diverse and high-quality <question, logical\_expression> pairs that involve unseen KB knowledge. 
Algorithm~\ref{alg:alg1} in Appendix~\ref{sec:algorithm} shows the organization process.

\vpara{Cluster-Based Diverse Logical Expressions Creation.} 
Relations in a logical expression serve as the foundation of the expression, reflecting the primary meaning of the logical expression. Therefore, the key to maintaining diversity in logical expressions lies in ensuring a diverse set of relations. 
Here we introduce (1) how to choose a diverse set of relations and combinations of relations, and (2) how to expand them to form logical expressions.

$\bullet$ \textbf{Relation Selection.} To construct one-hop logical expressions, we adopt all relations in the KB.
To construct multi-hop logical expressions, we design a cluster-based strategy to select combinations of relations instead of conducting exhaustive enumeration.
More clearly, we leverage BERT~\cite{devlin2018bert} to calculate an embedding for each relation by averaging the representations of the relation's tokens in the last hidden layer. Then we employ the classic K-Means~\cite{na2010research} algorithm to cluster relations based on their embeddings, where the number of clusters is set to be the number of classes in the KB ontology. After clustering, we proceed by randomly selecting a subset of relations from each cluster, creating a pool of candidate relations. Subsequently, from this pool, we sample multiple relations to construct a combination of relations.

$\bullet$ \textbf{Logical Expression Construction.}
For each relation or relation combination, we expand it to be the final logical expression by adding its connected classes, entities, and logical operators. 
Specifically, given a $KB_i$ with the ontology $\mathcal{O}_i=\{(\mathcal{C}_i \times \mathcal{R}_i \times \mathcal{C}_i)\}$, for a relation $r \in \mathcal{R}_i$, we first select a class $c_d$ that indicates the type of the answer entity, then we select an entity $e$ serve as the main entity in the logical expression. To ensure the executable of the logical expression, class $c_d$ is chosen from domain class set of $r$, satisfying $(c_d, r, c_r) \in \mathcal{O}_i$, and the entity $e$ is chosen from the range entity set $E{r} = \{e \in \mathcal{E}_i | \exists x \in \mathcal{E}_i, (x, r, e) \in \mathcal{F}_i\}$.
If the type of $e$ is literal, we select a logical function from the value operator set 
$\{$ARGMAX, ARGMIN, GT, GE, LT, LE, EQ$\}$, 
otherwise, we select from the entity operator set $\{$COUNT, AND, TOP,NONE$\}$. 
Combining these components (i.e., relation, class, entity, and logical operator), we expand the relation $r$ into a one-hop logical expression $s$. For complex logical expressions with multiple relations, we construct them by incrementally generating one-hop logical expressions and concatenating them. For example, given two relations $(r_1, r_2)$ where $r_1$ is connected to $r_2$ in the KB, we first construct logical expression $s_1$ and $s_2$ for $r_1$ and $r_2$, respectively. Then we append $s_2$ to $s_1$ to derive a multi-hop logical expression.

\vpara{Question Generation with Diverse Demonstrations.}
With the created logical expressions, we devise a KB-constrained prompt to instruct GPT-3.5 to generate natural language questions. The prompt consists of an instruction and a logical expression.
The human-written instruction is ``Given only the information below, following the examples, ask a factual question based on the given schema hint''.
The logical expression is described by a hint ``Relation: $T(r)$;  Class: $T(e)$; Entity: $T(e)$; Logical Operator: $T(g)$'',
where $T(\cdot)$ denotes the surface text of each schema which either be its name or description.
For each prompt, the in-context learning (ICL) demonstrations are placed before the hint.

Beyond traditional approaches such as random demonstration selection or uniform use of the same demonstrations across all target prompts, we introduce a diverse demonstration selection strategy to encourage GPT-3.5 to generate more diverse questions.
To implement this strategy, we first construct a question pool with questions sampled from existing general KBQA benchmark datasets, and then leverage BERT to embed the questions in the pool and the target prompt hint, separately.
In order to ensure the diversity of demonstrations, we employ the classic K-Means algorithm to cluster questions based on their embeddings. The number of clusters is set to be the number of classes in the KB ontology. 
Afterwards, we calculate cosine similarities between the embedding of the target prompt hint and each cluster center. 
We select $k$ questions from the most relevant cluster, ensuring they have distinct sentence structures. This is accomplished by clustering again the questions within the cluster into $k$ groups and randomly choosing one question from each group.
The selected questions along with their corresponding logical expressions, form the relevant demonstrations of the target prompt.
Notably, we mask entity mentions in each question to allow the clustering algorithm concentrate on the semantics of ontology-related information rather than specific instances.

\vpara{Reverse Checking.}
\label{sec:checking}
Considering the potential incorrect generation of GPT-3.5, we employ a reverse checking strategy to filter out the poor-quality generated questions. 
Given a question and logical\_expression pair <$q, s$>, we first convert $s$ which is in the form of s-expression into SPARQL and execute it to obtain the answer set $\mathcal{A}$ from KB. 
Then, we ask GPT-3.5 to judge whether the obtained answers $\mathcal{A}$ align with the generated question $q$ using the following prompt:
``Is $\mathcal{A}$ the answer set for question $q$? Please answer with YES or NO.''  It's important to note that in cases where the answer set contains extensive answers, we randomly sample 5 answers for checking. Any pair that receives a ``NO'' response is discarded.

Additionally, some of the generated questions are fluent in natural language but may not frequent asked by users. A typical example is, ``Who has the same gender with Obama's daughter?''. To filter out such atypical questions, we leverage GPT-3.5 to assess each question's likelihood of being posed by a user, by prompting: ``Score 1-10, where 1 signifies that the question is unlikely to be asked, and 10 implies that the question is frequently asked. Please score the question: $q$''. Questions receiving a score lower than 3 are considered improbable for user inquiry. Consequently, we exclude the training pairs about these questions from our augmented data.

\subsection{Knowledge Learning}

\vpara{Knowledge Refinement}. 
As the main target of LLaMA2 is to comprehend natural languages rather than logical expressions, we refine each response within the training data by adding the name of the queried KB and providing a natural language explanation of the logical expression, extending the training pairs in the form of <question, logical\_expression> to be pairs of <question, KB\_name; explanation; logical\_expression>.
The explanation encompasses relationships, classes, and entities relevant to the query, along with a description of the logical expression.
Given that entity IDs lack semantic information as numerical values, we prompt KBLLaMA to generate entity mentions instead. These mentions are in natural language format, enhancing the learning process. Consequently, we replace entity IDs with their corresponding mentions in logical expressions within the training data.
Furthermore, we translate logical operators into more easily understandable descriptions, such as replacing ARGMAX with MAX. This adaptation allows \smodel to effectively handle multiple KBs with distinct ontologies.



\vpara{Model Training}. With the organized knowledge, we fine-tune the LLaMA2-7B to take a natural language question as input and then output the corresponding logical expression with explanation.
Since entity mentions instead of IDs are output, we train an entity linker to identify the entity ID for each generated entity mention.
Given a question $q$ and the corresponding output logical expression by \model, we extract entity mentions from the logical expression and search for candidate entities whose names or aliases have token overlap with the mentions. 
We concatenate the question $q$ and the description of each candidate entity $e$ as <\text{question}: $q$; \text{entity}: $e$> and feed it into the pre-trained LLaMA2 (7B) to obtain the normalized logits $z$ in the last dense layer. Similar to T5-rank~\cite{2020t5}, the logit of of a special token ``<extra\_id\_10>'' is used for classifying whether the candidate entity aligns with the question. The ground truth label of the entity ID is obtained from our augmented <question, logical\_expression> pairs, where we can get the ground truth entity ID from the logical expression.
We adopt the  the classic classification objective --- cross-entropy, for training the entity linker.



For KBQA on general KBs, we can directly employ the \model-base, which is trained on existing general KBQA benchmark datasets and augmented data from general KBs. Concretely, we leverage 300,000 samples from general benchmark datasets, including GrailQA, WebQSP, CWQ on Freebase, and KQAPro on Wiki. 
Additionally, we generate 200,000 data samples based on Freebase and Wikidata to augment the training data. 
In this way, \model-base captures general knowledge for In-KB and Cross-KB generalization.
Additionally, users can customize \smodel for their domain-specific KBs via continuously fine-tuning the \model-Base with knowledge augmented from the domain-specific KBs, thereby enhancing the Cross-KB generalization capability. Empirically, we generate 100 training data instances for each  relation in the new KB. 
\section{Experiment}

\subsection{Experimental Setup}
\textbf{KBs and Datasets.}
We evaluate our methods across diverse KBs, including general large KBs like Freebase~\cite{freebase08} and Wikidata~\cite{pellissier2016wikidata}, as well as three smaller domain-specific KBs: Bio-Chemical~\cite{kushida2023exploring}, Academic~\cite{farber2019microsoft}, and Movie~\cite{tapaswi2016movieqa}. Details of each KB and their corresponding datasets are provided below.\\
$\bullet$ \textbf{Freebase Common Subset} comprises publicly available relations in Freebase~\cite{freebase08}, including 86 domains, 2,038 classes, 6,265 relations, and over 45 million entities. Most popular KBQA benchmarks are based on Freebase. GrailQA~\cite{gu2021beyond} focuses on generalization, with up to 4-hop logical expressions and complex operations. GrailQA consists of three sub-datasets: GrailQA (I.I.D), GrailQA (Zero), and GrailQA (Comp.). In GrailQA (I.I.D), both training and test sets share the same data distribution. GrailQA (Zero) presents relations and classes in the test set not seen during training. In GrailQA (Comp.), the test set introduces new compositions of previously seen relations and classes from the training set.
Besides, WebQSP~\cite{talmor-berant-2018-web} is an I.I.D. benchmark derived based on Freebase. CWQ is derived from WebQSP, featuring complex compositional queries.\\
$\bullet$ \textbf{Wikidata} contains structured knowledge from Wikipedia, with over 12 billion facts, 100+ million entities, and 10,000 relations. We focus on relations relevant for addressing natural language queries.
KQAPro~\cite{cao2022kqa} is a KBQA dataset that contains complex questions based on Wikidata, where the entities is extracted from Freebase. \\
$\bullet$ \textbf{Bio-Chemical KB} includes Bgee (gene expression)~\cite{bastian2021bgee} and OMA (orthology)~\cite{altenhoff2021oma} RDF stores, totaling 43,000 facts across 37 classes. Derived from it, the benchmark Bioinformatics~\cite{kushida2023exploring} encompasses complex biomedical queries, involving a series of intricate SPARQL queries. QALD-4~\cite{unger2014question} comprises 50 natural language biomedical questions, involving SPARQL queries from SIDER, Drugbank, and Diseasome domains. \\
$\bullet$ \textbf{Academic KB} is based on the Microsoft Academic Knowledge Graph (MAKG)~\cite{farber2019microsoft}, containing over 8 billion triples on scientific publications and entities. Due to the absence of KBQA benchmarks related the MAKG, we collect 100 question patterns from other Academic datasets~\cite{pollacci2022emakg} and replace concrete entities for creating questions and answers.\\
$\bullet$ \textbf{Movie KB} is a compact movie knowledge base extracted from OMDb, comprising 134,741 facts with nearly 4,000 entities. MQA~\cite{tapaswi2016movieqa} is based on this Movie KB, divided into three sub-datasets: MQA 1-hop, MQA 2-hop, and MQA 3-hop, based on the number of hops between answer entities and topic entities.

General large KB benchmarks have provided the training and test sets.
Domain-specific KB benchmarks are all treated as test sets. For benchmarks containing SPARQL queries, we convert them into s-expressions.

\vpara{Baselines.} 
We choose methods with competitive performance and high-quality open source code as our baselines.
The baselines can be divided into retrieve-the-reason models and prompting-based models.
In specific, the retrieve-the-reason models include RNG-KBQA (T5-based)~\cite{ye2022rng}, GrailQA-Rank (BERT-based)~\cite{gu2021beyond}, FC-KBQA (T5-based)~\cite{zhang2023fc-kbqa}, Pangu (T5-based)~\cite{gu-etal-2023-dont}, and ChatKBQA (LLaMA2-based)~\cite{luo2023chatkbqa}.
They are mainly proposed to solve the general KBs' QA tasks and enhance the In-KB generalization capability.
The prompting-based models leverage LLMs to generate the logical expressions. We consider GPT-3.5~\cite{tan2023chatgpt}, GPT-4~\cite{gpt4}, and LLaMA2 (13B) in the following experiments.
Additionally, we include comparisons with StructGPT~\cite{jiang-etal-2023-structgpt}, which employs sophisticated prompts for KBQA on LLMs. We employ the unified prompt ``Please answer the question:'' for GPT series and LLaMa2 (13B), while follow StructGPT's specific prompt for its implementation.


\vpara{Metrics.}
Exact Match (EM) and F1 are popular metrics for KBQA approaches. 
As some datasets do not include SPARQL queries as the ground truth, we do not evaluate the EM score where whether the generated logical expression exactly matches the ground truth. We only use F1 to measure the overlap between the predicted answer set and the gold answer set.

\vpara{Settings.}
For the base model, \textbf{\model-base} is trained with existing  benchmark datasets on general KBs and our augmented data on general KBs. Specifically, we train \smodel on 300,000 instances from benchmark datasets on Freebase and Wikidata, including GrailQA, WebQSP, CWQ on Freebase, and KQAPro on Wikidata, along with 200,000 augmented data derived from Freebase and Wikidata using the method proposed in Section~\ref{sec:approach}. We also evaluate a variant of the base model, \textbf{\model-base w/o G}, trained solely on the general KBs' benchmark dataset without any augmented data. 
\textbf{\smodel zero-shot}, \textbf{\smodel 5-shot}, and \textbf{\smodel 20-shot} are fine-tuned versions of the \model-base using augmented data on new KBs. Here zero/5/20-shot indicates the number of demonstrations used for constructing the <question, logical\_expression> training pairs.
Additional details about environment and hyper-parameters are provided in Appendix~\ref{sec:parameters}.

\begin{table*}[]
    \centering
    \renewcommand\arraystretch{1}
    \caption{Overall F1 performance on general KB's benchmark datasets (\%). }
    \label{tb:overall_performance}
    \vspace{-1em}
	\begin{tabular}{l|cccccc|cc}
	\toprule
    & \multicolumn{6}{c|}{Freebase} & \multicolumn{1}{c}{WikiData} \\ \cmidrule{2-8}
    & WebQSP & CWQ & GrailQA(All) & GrailQA(I.I.D) & GrailQA(Zero) & GrailQA(Comp.) & KQAPro \\ \midrule
	RNG-KBQA & 75.6 & 42.3 & 74.4 & 89.0 & 69.2 & 71.2 & - \\  
	GrailQA-Rank & 67.0 & - & 58.0 & 67.0 & 55.7 & 53.9 & 67.3 \\   
	FC-KBQA & 76.9 & 53.1 & 78.7 & \underline{91.2} & 74.0 & 76.7 & 79.0 \\
	Pangu & 79.6 & - & \underline{81.7} & 88.8 & \underline{78.5} & \underline{81.5} & - \\
	ChatKBQA  & 79.8 & \textbf{77.8}& 48.1 & 70.3 & 36.8 & 50.8 & 90.7\\ \midrule
	StructGPT & 72.6 & - & 45.7 & 38.2 & 54.1 & 33.3  & - \\
 LLaMA2 (13B) & 57.8 & 44.5 & 29.1 & 63.2 & 14.2 & 28.2 & 37.3 \\
	GPT-3.5 & 61.2 & 64.0 & 46.8 & 63.9 & 46.0 & 39.3 & 47.9 \\ 

 GPT-4 & \textbf{90.5} & 71.0 & 51.4 & 57.3 & 52.2 & 48.4 & 57.2 \\ \midrule   
	\model-base w/o G & 79.3 & \underline{76.7} 
 & 46.5 & 63.6 & 39.0  & 45.7
  & \underline{92.1} \\ 
        \model-base & \underline{82.3} & 76.3 & \textbf{84.4} & \textbf{94.5}  & \textbf{85.0} & \textbf{84.8} & \textbf{94.0} \\
 \bottomrule
\end{tabular}
\end{table*}

\subsection{Evaluation on General KBs}
We evaluate \model's performance on general KBs' benchmark datasets, including both I.I.D. and In-KB generalization settings. Results are summarized in Table~\ref{tb:overall_performance}, with bold indicating the best and underline indicating the second-best performances. Overall, our model consistently outperforms baseline models across the majority of KBQA benchmarks, demonstrating its efficacy in leveraging and reasoning over knowledge within general KBs. This positions our model as a robust foundation for future KBQA learning endeavors. Detailed findings are provided below.

\textbf{(1) \model-base outperforms all the retrieve-based models on most datasets}, including RNG-KBQA (T5-based), GrailQA (BERT-based), FC-KBQA (T5-based), Pangu (T5-based), and ChatKBQA (LLaMA2 7B-based). 
\smodel achieves promising results within a simple framework without additional components such as a retriever in the retrieve-and-reason framework. Notably, \smodel surpasses ChatKBQA, which also utilizes LLaMA2-7B, demonstrating the effectiveness of our methods beyond a strong base model and benefiting from high-quality augmented training data.

\textbf{(2) \model-base outperforms GPT-3.5 across all datasets and even surpasses GPT-4 on most datasets}. This underscores the importance of fine-tuning the model with training data to reveal ontology information of the KB. Simply prompting GPT-3.5 or GPT-4 does not suffice to recognize such information. Additionally, we outperform StructGPT, which designs sophisticated prompts for GPT-3.5 but struggles with multi-hop questions containing intricate logical operators like ARGMAX.

\textbf{(3) \model-base outperforms all models specifically tailored to enhance In-KB generalization}.
GrailQA (Zero) and GrailQA (Comp.) correspond to the In-KB generalization scenarios.
Previous retriever-based model designed for the In-KB generalization heavily rely on semantic similarity between logical expressions and questions, lacking the comprehensive training data encompassing all KB relations for zero-shot generalization achieved by our approach. Also, our data construction method incorporates intricate functions and multi-hop relations, empowering our model to adeptly handle compositional generalization.


\textbf{(4) Removing augmented data decreases accuracy}. Removing our augmented data results in accuracy drops on most datasets, indicating the significant role of our augmented data in enhancing performance. 
The notable accuracy drop is due to the complex and diverse nature of test data in some datasets like GrailQA, where sufficient training data notably enhances model's capabilities. Conversely, in datasets like KQAPro, the simplicity of test data and adequacy of training data in the benchmark diminishes the impact of excluding our augmented data.

\subsection{Evaluation on Cross-KB Generalization}
\begin{table}[]
    \centering
    \renewcommand\arraystretch{1}
    \caption{Cross-KB Generalization performance (F1 \%).}
    \label{tb:transfer_performance}
    \vspace{-1em}
    \begin{tabular}{l|ccccc}
    \toprule
    & MQA & MQA & MQA & Bio. & Aca. \\
    & 1-hop & 2-hop & 3-hop & & \\\midrule  
    FC-KBQA & 92.1 & 89.2 & 83.4 & 80.8 & 78.5 \\
    Pangu & 94.5 & 90.3 & 89.2 & 78.2 & 80.9\\ 
    ChatKBQA & 28.1 & 16.7& 10.4 & 45.8 & 30.7\\  \midrule
    StructGPT & \underline{97.1} & \textbf{97.3} & 87.0 & 59.0 & 68.9\\ 
    LLaMA2 (13B) & 40.5 & 32.6 & 20.1 & 33.4 & 14.5 \\
    GPT-3.5 & 61.9 & 31.0 & 43.2 & 54.9 & 34.2\\  
    GPT-4 & 69.6  & 75.5 & 59.2 & 68.9 & 58.0\\  \midrule   
    \smodel zero-shot & 87.3 & 76.9 & 70.8 & 85.0 & 81.6\\
    \smodel  5-shot & 90.9 & 84.5 &  \underline{89.3} & \underline{88.6} & \underline{84.0}\\ 
    \smodel 20-shot & \textbf{98.7} & \underline{96.1}  & \textbf{97.2} & \textbf{91.6} & \textbf{86.2} \\ 
     \bottomrule
\end{tabular}
\end{table}

To evaluate the Cross-KB transfer performance, we conduct evaluations on three new KBs: Movie KB , Bio-Chemical KB, and Academic KB.
A portion of the knowledge within Movie KB might be encompassed by the more general Freebase and Wikidata. Consequently, \model-base has captured some knowledge in Movie KB.
Specially, Bio-Chemical KB and Academic KB are two new domain-specific KBs posing greater challenges due to most of the unseen ontologies and domain knowledge. 


For each new KB, we employ our data augmentation process to generate 100 training instances for each relation, incorporating diverse entities. These instances are used to fine-tune \model-base. We select the best-performed fine-tuned baselines from Table~\ref{tb:overall_performance} for comparison and present the comparison results in Table~\ref{tb:transfer_performance}. 

\textbf{(1) \smodel consistently delivers promising results across all new KBs.}
Raw LLMs including GPT-3.5, GPT-4, and LLaMA2 (13B) perform poorly, especially on datasets requiring highly domain-specific knowledge, such as identifying authors of low-cited papers. While smaller models like FC-KBQA and Pangu show acceptable performance on MQA which is a KBQA dataset based on Movie KB, their performance decreases on Bio-Chemical and Academic datasets because their retrievers lack specific domain knowledge.

\textbf{(2) \smodel demonstrates promising performance in handling generalization on multi-hop questions.}
For MQA, we present results for each hop in the first three columns of Table~\ref{tb:transfer_performance}. \smodel is fine-tuned with newly augmented training data for movies, covering up to 4,000 multi-hop training pairs. Since our augmented data covers all one-hop relations, \smodel shows promising performance on one-hop questions. Although the augmented data does not include all multi-hop questions, we achieve favorable performance in multi-hop settings. This highlights the model's capacity to learn and retain the semantics of relations through one-hop training and its ability to reason through multi-hop questions.

\textbf{(3) Few-shot demonstrations enhance \model's performance.}
We compare our model's performance in zero-shot and few-shot scenarios. In zero-shot situations, 
the GPT-3.5 generates questions solely based on the created logical expressions. In the few-shot setting, we incorporate diverse demonstration examples to enhance the generation quality of GPT-3.5. 
Across all the three new KB datasets, the highest performance is achieved with 20-shot setting. As the number of shots decreases, there is a drop in performance, though the extent of this decline varies across different domains. Academic datasets exhibit a smaller decline compared to Bio-Chemical, likely owing to their general nature, which closely aligns with the LLM's familiarity and results in higher-quality augmented questions. Conversely, in less familiar domains such as Bio-Chemical, the model benefits more from few-shot guidance.

\subsection{Ablation Study}

\vpara{Effect of Augmented Data Amount.}
\begin{figure}[t]
	\centering
	\subfigure[\model-base]{\label{subfig:datasize}
		\includegraphics[width=0.23\textwidth]{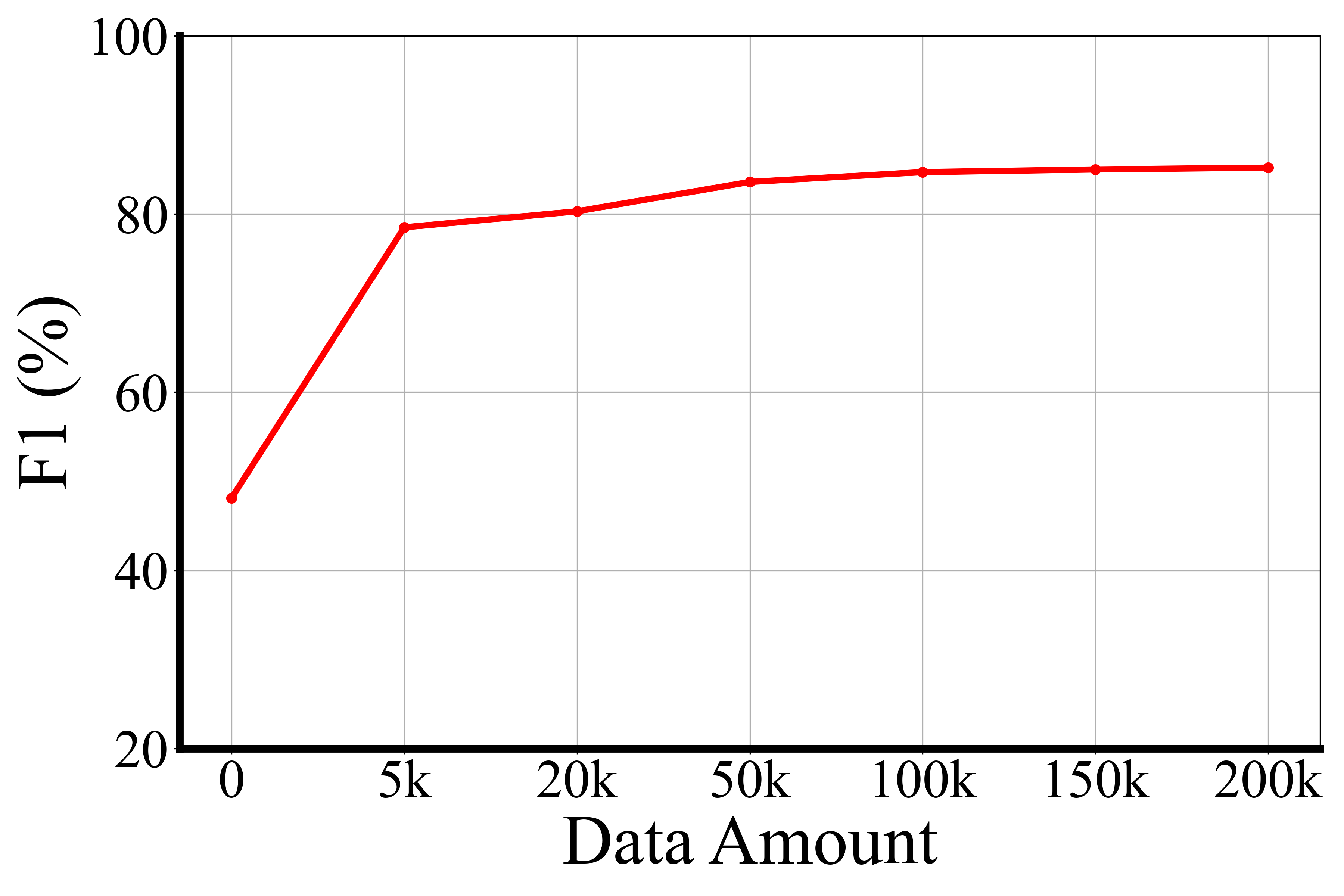}
	}
        \hspace{-0.2cm}
	\subfigure[\model-base Transfer]{\label{subfig:dataratio}
		\includegraphics[width=0.23\textwidth]{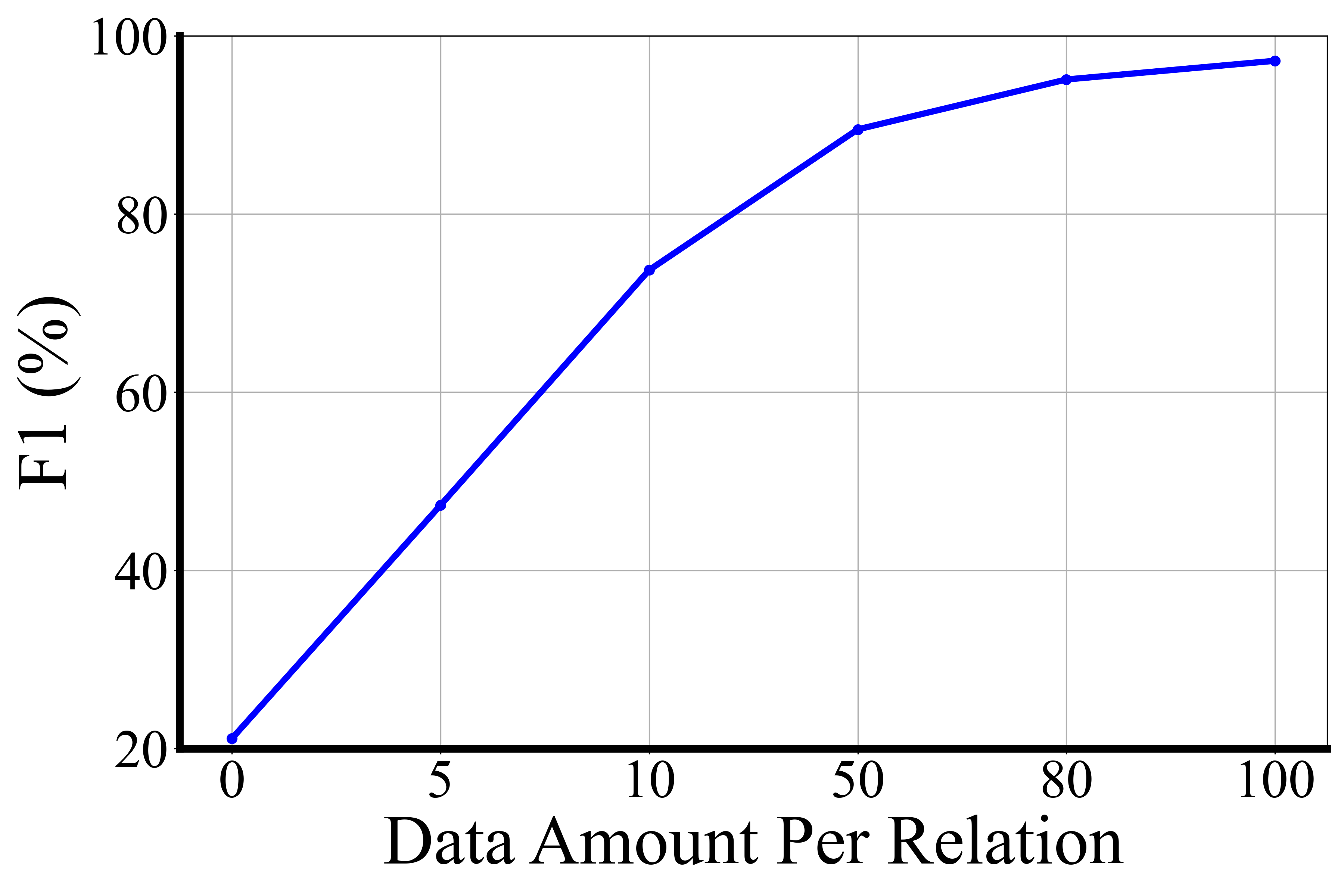}
	}		
 \caption{\label{exp_fig} Effect of augmented data amount. We evaluate the performance of \model-base on GrailQA (All) and the transfer performance of \model-based on MQA.}
\end{figure}
Considering the high cost of API calls for data augmentation, we conduct an ablation study to determine the amount of the augmented data for achieving satisfactory model performance. Concretely, we explore two points: the data amount required to train the \model-base and the data amount needed for fine-tuning the \model-base to transfer to a new KB. 
We evaluate \model-base on GrailQA(All) and evaluate a fine-tuned version of \model-base on MQA under the 20-shot setting.

The results are depicted on the left and right sides of Figure~\ref{exp_fig}, respectively.
The results suggest that increasing the amount of augmented data contributes to enhancing model performance, however, as the amount reaches a certain threshold, the increase rate tends to diminish.
Besides, we observe that the increase of performance on GrailQA is less notable than that on MQA. That is because the training and test sets of GrailQA are based on Freebase, and the augmented data for training \model-base also come from the Freebase. Hence increasing the augmented data amount does not bring a significant performance gain.
In contrast, MQA is based on the domain-specific Movie KB. Hence \model-base is less familiar with the Movie KB compared with Freebase. Therefore, increasing the augmented data amount will bring more notable performance gain.




\vpara{Effect of Knowledge Organization and Refinement Strategy.}
\begin{table}[]
    \centering
    \renewcommand\arraystretch{1}
    \caption{Effect of knowledge organization and refinement.}
    \label{tb:ablation}
    \vspace{-1em}
    \begin{tabular}{l|c}
    \toprule
    & Bio-Chemical (F1 \%) \\ \midrule  
    \smodel 20-shot & 91.6 \\
    w/o diverse relation selection  & 90.6 \\  
    w/o demonstrations & 89.8 \\ 
    w/o reverse checking & 90.8  \\ 
    w/o knowledge refinement & 87.4 \\
    \bottomrule

\end{tabular}
\end{table}
We explore the impact of knowledge organization and refinement under the few-shot transfer setting, where \model-base is transferred to the Bio-Chemical KB with 20-shot setting. As illustrated in Table~\ref{tb:ablation}, removing diverse relation selection (replace with random selection), demonstrations in in-context learning, or the reverse checking strategy for ensuring data quality will lead to a decrease in F1 score. This observation indicates the importance of ensuring diversity and high-quality of the augmented training data. It is worth noting that the decrease in F1 score is not obvious, as Bio-Chemical has a limited number of relations. Even using random selection, it is possible to cover all relations and maintain diversity.
Furthermore, we demonstrate the effectiveness of knowledge refinement. Removing the knowledge refinement strategy results in a 4.2\% drop in F1 score. This indicates that the knowledge refinement inspired by the idea of chain-of-thought can facilitate the learning of \model, thereby enhancing the overall performance.

\subsection{Evaluation on Entity Linker}
\begin{table}[]
    \centering
    \renewcommand\arraystretch{1}
    \caption{Entity linking accuracy on GrailQA (\%).}
    \label{tb:entity_acc}
    \vspace{-1em}
    \begin{tabular}{l|c}
    \toprule
    & Accuracy\\ \midrule  
    BERT-based Entity Linker & 72.2 \\
    RNG-KBQA's Entity Linker & 81.6 \\ 
    FC-KBQA's Entity Linker & 87.2 \\  
    Ours's Entity Linker & \textbf{91.3}  \\ \bottomrule
\end{tabular}
\end{table}
We compare our entity linker with existing entity linkers in Table~\ref{tb:entity_acc}. BERT-based entity linker identifies mentions using a BERT-based name entity recognition (NER) model and then select entities according to their frequency.
RNG-KBQA employs another BERT model for ranking entities. FC-KBQA uses an entity pruning strategy to remove irrelevant entities.

Our entity linker surpasses all these baselines, highlighting the effectiveness of training the LLaMA2-7B as an entity linker for KBQA tasks. Our entity linker determines whether an entity description aligns with the question, even without KB constraint pruning. In contrast to baselines relying on an additional mention detector, our approach directly extracts mentions from \model's output logical expressions, achieving better performance and reducing the need for additional components.
\section{Conclusion}
This paper addresses generalization challenges in KBQA, with the goal of enhancing the model's adaptability in both In-KB and Cross-KB scenarios. Whether tackling In-KB or Cross-KB generalization, the crucial factor is the introduction of previously unseen knowledge.
Current approaches predominantly adopt the retrieve-then-reason framework, often overlooking the significance of incorporating new knowledge into the model parameters, thereby limiting their generalization capabilities.
As a solution, this paper introduces a learn-then-reason model named \smodel, designed to inject new KB knowledge into the widely-used open-source LLaMA2 (7B) for a flexible end-to-end KBQA. \smodel demonstrates promising performance on KBQA generalization, particularly in Cross-KB tasks which are less studied by prior work, contributing to addressing the domain-specific KBQA tasks.


\bibliographystyle{ACM-Reference-Format}
\bibliography{custom}

\clearpage
\appendix
\section{Appendix}\label{sec:appendix}
\startcontents[appendix]
\printcontents[appendix]{ }{1}{\setcounter{tocdepth}{2}\vspace{5pt}}

\subsection{S-Expression}
\label{sec:appendix_a}
The s-expression integrates a variety of function operators and exhibits strong compositional capabilities. This format encompasses relations, classes, entities, and logical operators, with the class specifying the domain class of the answer. The \textbf{logical operators} include a set of projection operators {``LE'', ``LT'', ``GT'', ``GE'', ``EQ'', ``JOIN''} and a set of function operators {``AND'', ``COUNT'', ``ARGMIN'', ``ARGMAX''}. Table~\ref{tab:logical_operator} shows the logical operators used in s-expression, where $r$ denotes a relation, $e$ denotes an entity, and $E_1, E_2$ denote two sets of entities. 

\subsection{Logical Expression Creation Algorithm}\label{sec:algorithm}
The details of one-hop logical expression construction are shown in algorithm~\ref{alg:alg1}, where we input a new KB and the question corpus collected from the benchmark, and then output the questions and logical expressions pairs based on the new KB. For the multi-hop logical expressions, we first enumerate the connected relation set and then construct logical expressions for each relation in the relations set, finally we concatenate these logical expressions together to form the multi-hop logical expression. For example, given a relation ``music.album.artist'' and a relation ``game.video\_game. soundtrack'' which are connected in KB, we first construct a logical expression for each of them, ``AND music.album (JOIN music.album.artist Gustavo\_Santaolalla )'', ``AND game.video\_game (JOIN game. video\_game. soundtrack All\_Gone)'', and finally, we concatenate these two logical expressions together into ``AND game. video\_game (JOIN game. video\_game. soundtrack (JOIN music. album. artist Gustavo\_Santaolalla))'' and form a two-hop logical expression.

\begin{table}
  \caption{Definition of logical operators in s-expression.}
  \label{tab:logical_operator}
  \begin{tabular}{cl}
    \toprule
    Operator  & Definition\\
    \midrule
    JOIN & \makecell[l]{(JOIN $r$ $e$) denotes querying of the head entity \\in a triple (?, $r$, $e$)}\\
    GE  & \makecell[l]{(GE $E_1$ $e$) denotes a subset of $E_1$ which the literal \\in it is greater than or equal to literal $e$}\\
    GT  & \makecell[l]{(GT $E_1$ $e$) denotes a subset of $E_1$ which the literal \\in it is greater than literal $e$}\\
    LE  & \makecell[l]{(LE $E_1$ $e$) denotes the subset of $\mathcal{E}_1$ which the literal \\in it is less than or equal to literal $e$}\\
    LT  & \makecell[l]{(LT $E_1$ $e$) denotes the subset of $\mathcal{E}_1$ which the literal \\in it is less than literal $e$}\\
    EQ  & \makecell[l]{(EQ $E_1$ $E_2$) denotes whether the entity set $E_1$ \\is equals to the entity set $E_2$}\\
    AND   & \makecell[l]{(AND $E_1$ $E_2$) denotes the intersection of entity \\set $E_1$ and entity set $E_2$}\\
    COUNT  & \makecell[l]{(COUNT $E_1$) denotes the size of the entity set $E_1$}\\
    ARGMAX  & \makecell[l]{(ARGMAX $E_1$ $r$) denotes the maximal literal \\obtained after (JOIN $r$ $E_1$) = \{(JOIN $r$ $e$)|$e \in E_1$\}}\\
    ARGMIN  & \makecell[l]{(ARGMIN $E_1$ $r$) denotes minimal literal obtained \\after (JOIN $r$ $E_1$) = \{(JOIN $r$ $e$)|$e \in E_1$\}}\\
    \bottomrule
  \end{tabular}
\end{table}

\begin{algorithm}
 \renewcommand{\algorithmicrequire}{\textbf{Input:}}
	\renewcommand{\algorithmicensure}{\textbf{Output:}}
	\renewcommand{\algorithmicreturn}{\textbf{Return}}
	\caption{\textbf{Diverse Logical Expressions Creation}}
	\label{alg:alg1}
	\begin{algorithmic}[1]
	    \REQUIRE A knowledge base (consists of relation set $\mathcal{R}$, entity set $\mathbf{E}$, class set $\mathbf{C}$, ontology set 
 $\mathbf{O}$, and relational fact set $\mathbf{F}$ ), the large language model GPT-3.5, and the question corpus $\mathcal{Q}_b$ from benchmark dataset.
	    \ENSURE Question and logical expression pairs $\mathcal{P}_g=\{(q_i, s_i)\}$. 
            \STATE Initialize the $\mathcal{P}_g = \varnothing$.
	    \FOR{$r$ in $\mathbf{R}$}
            \STATE Randomly select a subset of connected classes $C_r$;
            \STATE Randomly select a subset of connected entities $E_r$;
            \FOR{$e$ in $E_r$}
                \STATE Initialize the logical operator $g$ = NONE.
                \IF{$e$ is literal}
                    \STATE Randomly select from \{MAX, MIN, GT, LT, GT, LE, GE, EQ, NONE\} and assign $g$;
                \ELSE
                    \STATE Randomly select  from \{COUNT, AND\} and assign $g$; 
                \ENDIF
	       \FOR{$c$ in $C_r$}
                \STATE Construct the logical expression $s_i$ based on $(r,e,c,o)$;
                \STATE Construct prompt hint $h_i$ based on $(r,e,c,o)$;
                \STATE Select relevant question set $Q_q$ from $\mathcal{Q}$, based on $h_i$;
                \STATE Cluster $Q_q$ into $\{Q_1,Q_2,...\}$ according to its pattern;
                \FOR{each cluster set $Q_i$ in  $Q_q$}
                    \STATE Randomly select a question $q_e$ from $Q_i$;
                    \STATE Prompt GPT-3.5 with $q_e$, $h_i$ to generate question $q_i$;
                    \IF{Reverse-Check($q_i, s_i$)}
                        \STATE $\mathcal{P}_g$ = $\mathcal{P}_g \cup \{(q_i,s_i)\}$;
                    \ENDIF
                \ENDFOR
	       \ENDFOR
	    \ENDFOR
     \ENDFOR
	    \STATE Return question and logical expression pairs $\mathcal{P}_g$;
	\end{algorithmic}  
\end{algorithm}

\subsection{Experimental Settings}
\label{sec:parameters}

All the experiments are written in the Pytorch framework and run on 4 GPUs with 48GB of memory. For large general KBs, we build a virtuoso database following GrailQA which needs at least 100 GB RAM for the service. For small domain-specific KB, we directly read data into the service's memory while programming.  We call GPT-3.5 through the open official APIs. To train the s-expression generation model and the entity linker, we fine-tune LLaMa2 (7B) for 3 epochs with a batch size of 128, a learning rate of $4e^{-6}$, and a max context length of 4,096. The learning rate has a cosine decay and a linear warmup for the initial 5\% of training.
\end{document}